\documentclass[conference]{IEEEtran}
\IEEEoverridecommandlockouts

\usepackage{cite}
\usepackage{amsmath,amssymb,amsfonts}
\usepackage{algorithmic}
\usepackage{graphicx}
\usepackage{textcomp}
\usepackage{xcolor}
\usepackage{hyperref}
\def\BibTeX{{\rm B\kern-.05em{\sc i\kern-.025em b}\kern-.08em
    T\kern-.1667em\lower.7ex\hbox{E}\kern-.125emX}}
\newcommand\norm[1]{\left\lVert{#1}\right\rVert}
\newcommand{\myvec}[1]{\mbox{\boldmath $#1$}}

\newcommand\blfootnote[1]{%
  \begin{NoHyper}%
  \renewcommand\thefootnote{}\footnote{#1}%
  \addtocounter{footnote}{-1}%
  \end{NoHyper}%
}

\begin{document}

\title{Conditional Deep Canonical Time Warping}

\author{
\IEEEauthorblockN{Ran Eisenberg*}
\IEEEauthorblockA{\textit{Faculty of Engineering} \\
\textit{Bar Ilan University}\\
Ramat Gan, Israel \\
eisenbr2@biu.ac.il}
\and
\IEEEauthorblockN{Afek Steinberg*}
\IEEEauthorblockA{\textit{Faculty of Engineering} \\
\textit{Bar Ilan University}\\
Ramat Gan, Israel \\
afeks98@outlook.com}
\and
\IEEEauthorblockN{Ofir Lindenbaum}
\IEEEauthorblockA{\textit{Faculty of Engineering} \\
\textit{Bar Ilan University}\\
Ramat Gan, Israel \\
ofirlin@gmail.com}
}

\maketitle

\begin{abstract}
Temporal alignment of sequences is a fundamental challenge in many applications, such as computer vision and bioinformatics, where local time shifting needs to be accounted for. Misalignment can lead to poor model generalization, especially in high-dimensional sequences. Existing methods often struggle with optimization when dealing with high-dimensional sparse data, falling into poor alignments. Feature selection is frequently used to enhance model performance for sparse data. However, a fixed set of selected features would not generally work for dynamically changing sequences and would need to be modified based on the state of the sequence. Therefore, modifying the selected feature based on contextual input would result in better alignment. Our suggested method, Conditional Deep Canonical Temporal Time Warping (CDCTW), is designed for temporal alignment in sparse temporal data to address these challenges. CDCTW enhances alignment accuracy for high dimensional time-dependent views by performing dynamic time warping on data embedded in maximally correlated subspace, which handles sparsity with a novel feature selection method. We validate the effectiveness of CDCTW through extensive experiments on various datasets, demonstrating superior performance over previous techniques.
\end{abstract}


\section{Introduction}
\label{sec:introduction}
Temporal alignment in time series data is crucial for many machine learning and signal processing applications. Traditional methods, such as Dynamic Time Warping (DTW) \cite{kruskal1983dtw} and Canonical Time Warping (CTW)  \cite{zhou2009canonical}, are susceptible to complex temporal and noisy data dependencies. While these techniques have been widely used, they often struggle with high-dimensional data. They are limited by their linear assumptions, mainly when dealing with sparse data where alignment errors can significantly impact model performance.

Recent advancements, such as Deep Canonical Time Warping (DCTW) \cite{trigeorgis2016deep}, have attempted to address these challenges by leveraging deep learning to capture non-linear relationships. However, DCTW and similar methods still face optimization difficulties, especially in the presence of sparse, high-dimensional data. This often leads to suboptimal alignments, which can degrade the performance of subsequent tasks.

Feature selection becomes increasingly important in the context of sparse, high-dimensional data, where irrelevant or redundant features can exacerbate the difficulties of temporal alignment. Stochastic gates (STG) \cite{pmlr-v119-yamada20a} is an embedded feature selection algorithm that incorporates feature selection directly into the model's optimization process by imposing regularization constraints on the loss function. This method has been demonstrated to enhance model performance both in supervised scenarios \cite{yang2022locally,jana2023support} and unsupervised settings \cite{svirskyinterpretable}. A recent extension, Conditional Stochastic Gates (c-STG) \cite{dyuthi2023contextual}, further advances this approach by modeling feature importance using conditional Bernoulli variables whose parameters are predicted based on contextual information. This technique allows for context-dependent feature selection, which improves flexibility and accuracy in complex feature selection tasks.

In this work, we introduce a novel conditional feature selection method into the Dynamic Time Warping (DTW) framework, addressing the limitations of traditional approaches in handling high-dimensional and sparse data. Specifically, we integrate Conditional Stochastic Gates (c-STG) into the alignment process, allowing the model to dynamically select the most relevant features based on the contextual information. This integration enhances the robustness of the alignment by reducing the influence of irrelevant or redundant features. By incorporating a conditional feature selection strategy, CDCTW offers a more flexible and accurate solution to the temporal alignment problem, particularly in high-dimensional settings.

Our new method, Conditional Deep Canonical Temporal Time Warping (CDCTW), includes a unique unsupervised feature selection scheme extending the $\ell_0$-based sparse Canonical Correlation Analysis (CCA) \cite{lindenbaum2021l0}. This allows the model to dynamically select relevant features during the alignment process, significantly improving alignment accuracy by reducing the input data's dimensionality and enhancing the robustness of the warping process. CDCTW achieves state-of-the-art performance on several datasets. We evaluate CDCTW on multiple datasets across various synthetic and benchmark datasets and demonstrate its superior performance in terms of alignment score compared to existing techniques such as Canonical-Soft DTW (CSTW/ASTW) \cite{kawano2019canonical}, DCTW \cite{trigeorgis2016deep}, Canonical DTW (CTW) \cite{zhou2009canonical}, and Generalized DTW \cite{deriso2022general}.

\section{Background}
\label{sec:background}


\subsection{Time Warping}
\label{sec:dtw}
Dynamic Time Warping (DTW) is an algorithm that temporally aligns two temporal sequences that may vary in speed or synchronization. It finds an optimal alignment between sequences by stretching or compressing the time axis, minimizing the distance measured between them. DTW is widely applied in various fields such as speech recognition \cite{muda2010voice}, gesture analysis \cite{jeong2011gesture}, and time series data mining \cite{rakthanmanon2012searching} due to its robustness in handling temporal variations.

Given two data sequences $\myvec{X}\in \mathbb{R}^{D\times N_x}$ and $\myvec{Y}\in \mathbb{R}^{D\times N_y}$, DTW attempts to temporally align by stretching and warping the time axis of each sequence in an optimal sense of some distance function $d(x, y)$ (most commonly $\ell_1$ or $\ell_2$ norm). The output of DTW are two alignment paths $P_x$ and $P_y$ that assign samples from the first sequence to relevant samples in the second sequence. These alignment paths must satisfy the continuity and monotonicity constraints. By noting the permuted sample of $\myvec{X}$ and $\myvec{Y}$ by the alignment paths as $\myvec{X}^{(P_x)}$ and $\myvec{Y}^{(P_y)}$, DTW’s objective can be written as
\begin{equation*}
    \min_{P_x, P_y} d\left(\myvec{X}^{(P_x)}, \myvec{Y}^{(P_y)}\right).
\end{equation*}

The number of alignment paths is exponential with respect to the number of samples of each view; however, this problem is solved using a dynamic programming algorithm that finds the optimal paths in $\mathcal{O}(N_xN_y)$.
A disadvantage of DTW is that it is restricted to sequences with a similar number of features. Hence, Canonical Time Warping (CTW) \cite{zhou2009canonical} has been suggested. This method shares the same goal as DTW but integrates dimensionality reduction of both views into a shared maximally correlated subspace by using CCA before aligning the data. CTW’s objective can be written as
\begin{equation*}
    \min_{P_x, P_y} d\left(\myvec{a}^T\myvec{X}^{(P_x)}, \myvec{b}^T\myvec{Y}^{(P_y)}\right),
\end{equation*}
where $\myvec{a}$ and \myvec{b} are the canonical vectors.
A drawback of CTW would be its linearity restriction; thus, Deep Canonical Time Warping (DCTW) \cite{trigeorgis2016deep} has been proposed, where the canonical vectors are replaced by NNs.

Soft Dynamic Time Warping (Soft-DTW) \cite{kawano2019canonical} was introduced to address DTW's sensitivity to local minima by smoothing the optimization landscape through a differentiable, soft-minimum approach. To improve alignment accuracy, annealed Soft-DTW (ACTW) gradually decreases a temperature parameter $\gamma$, transitioning from soft to hard alignments, making it more effective in applications requiring precise sequence alignment.

\subsection{Canonical Correlation Analysis}
\label{sec:cca}
Canonical correlation analysis (CCA) \cite{10.1093/biomet/28.3-4.321} is a classical statistical method focusing on finding linear projections of two measurements captured by coupled modalities or observation devices. CCA seeks linear transformations that maximize the correlation of the two projected data points \cite{6788402}.

Given the modalities $\myvec{X}\in \mathbb{R}^{D_x\times N}$ and $\myvec{Y}\in \mathbb{R}^{D_y\times N}$, the goal of CCA is to find projections $\myvec{a} \in \mathbb{R}^{D_x}$ and $\myvec{b} \in \mathbb{R}^{D_y}$ such that:
\fontsize{8.5pt}{8.5pt}\selectfont
\begin{equation*}
\begin{split}
    \max_{\myvec{a}, \myvec{b} \neq \mathbf{0}} \rho\left(\myvec{a}^\top \myvec{X}, \myvec{b}^\top \myvec{Y}\right) = \max_{\myvec{a}, \myvec{b} \neq \mathbf{0}} \frac{\myvec{a}^\top \myvec{C}_{xy} \myvec{b}}{\sqrt{\myvec{a}^\top \myvec{C}_{xx} \myvec{a} \myvec{b}^\top \myvec{C}_{yy} \myvec{b}}}.
\end{split}
\end{equation*}
\fontsize{10pt}{12pt}\selectfont
Where $\myvec{C}_{xy}$, $\myvec{C}_{xx}$, and $\myvec{C}_{yy}$ denote the covariance matrices of $\myvec{X}$ and $\myvec{Y}$, and $\rho$ is the correlation. $\myvec{a}, \myvec{b}$ are found using the solution of the  generalized eigenpair problem $\mathbf{C}_{xx}^{-1}\mathbf{C}_{xy}\mathbf{C}_{yy}^{-1}\mathbf{C}_{yx}\myvec{a}=\lambda^2\myvec{a} \;\;\Rightarrow\;\; \myvec{b}=\frac{\mathbf{C}_{yy}^{-1}\mathbf{C}_{yx}}{\lambda}\myvec{a}.$


\begin{figure*}[t!]
    \centering
    \includegraphics[width=\textwidth]{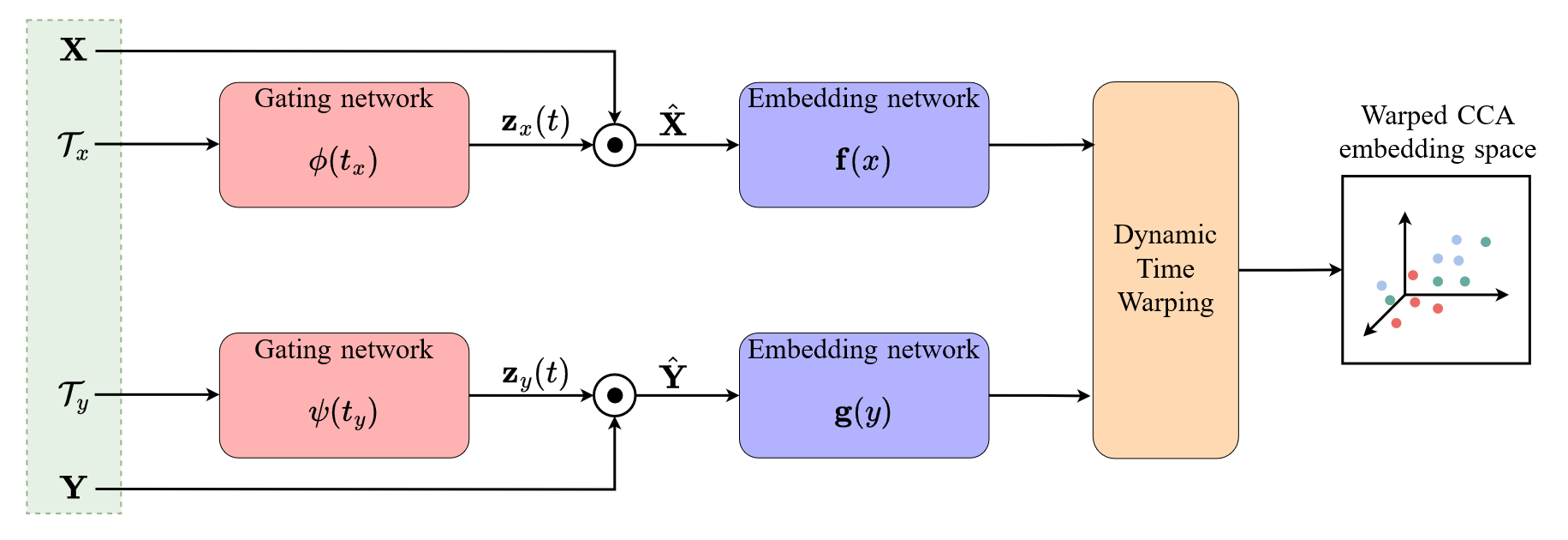}
    \caption{The proposed CDCTW architecture: Inputs $\mathcal{T}_x$ and $\mathcal{T}_y$ are plugged in to the conditional networks $\phi$ and $\psi$ correspondingly. Gate values $\myvec{z}_x$ and $\myvec{z}_y$ are then used to modify the inputs $\myvec{X}$ and $\myvec{Y}$ as described in section \ref{sec:method}. The modified input is fed into $\mathbf{f}$ and $\mathbf{g}$, which produce suitable embeddings for dynamic time wrapping.}
    \label{fig:arch}
\end{figure*}

CCA provides a simple framework with a closed-form solution. However, it is restricted to linear projections, which may not adequately capture non-linear relationships in many scenarios. Several non-linear extensions to CCA have been proposed \cite{bach2002kernel, michaeli2016nonparametric, lindenbaum2020multi,lindenbaum2015learning}, most notably Deep CCA (DCCA) \cite{andrew2013deep,lindenbaum2019seismic}.

DCCA attempts to learn non-linear projections using neural networks. Given modalities $\myvec{X}$ and $\myvec{Y}$, and neural network functions ${\mathbf{f_\theta}_x}: \mathbb{R}^{D_x}\rightarrow \mathbb{R}^o$, ${\mathbf{g_\theta}_y}: \mathbb{R}^{D_y}\rightarrow \mathbb{R}^o$, where $o$ is the embedding dimension. DCCA aims to find parameters $\{\boldsymbol{\theta}_x, \boldsymbol{\theta}_y\}$ such that
\begin{equation*}
     \min_{\myvec{\theta}_x, \myvec{\theta}_y} -\rho\left(\mathbf{f}(\myvec{X}; \myvec{\theta}_x), \mathbf{g}(\myvec{Y}; \myvec{\theta}_y)\right).
\end{equation*}
This allows DCCA to capture non-linear relations between modalities, making it more suitable than CCA in scenarios where linear assumptions do not hold \cite{qiu2018multi, sun2020learning}. The main challenge with both CCA and its non-linear extensions arises when $D_x \gg N$ or $D_y \gg N$, where accurate estimation of covariance matrices becomes impractical.

\subsection{$\ell_0$-based Sparse CCA}
\label{sec:l0cca}
The $\ell_0$-based Sparse Canonical Correlation Analysis ($\ell_0$-CCA) \cite{lindenbaum2021l0} addresses challenges encountered in Canonical Correlation Analysis (CCA) and Deep Canonical Correlation Analysis (DCCA) when dealing with high-dimensional data, particularly when the number of dimensions ($D$) greatly exceeds the number of samples ($N$). To achieve this, the method introduces an unsupervised embedded feature selection module, which employs a stochastic gating layer \cite{pmlr-v119-yamada20a} on the input data. The number of features used will effectively become smaller, allowing us to estimate the covariances more accurately and making CCA and DCCA feasible.
By noting the sets of relevant features of both views as $\mathcal{S}^x$ and $\mathcal{S}^y$ we can define the indicator vectors $\myvec{s}^x$ and $\myvec{s}^y$ such that $\forall d\in[1,D_x]: d\in \mathcal{S}^x\implies \myvec{s}^x_d=1$ and $\forall d\in[1,D_y]: d\in \mathcal{S}^y\implies \myvec{s}^y_d=1$. By using this notation, the objective of $\ell_0$-CCA can be written as
\vspace{-0.12cm}
\begin{equation*}
\begin{split}
    \min_{\myvec{a}, \myvec{b}} &-\rho \left(\left(\myvec{a}\odot\myvec{s}^x\right)^\top \myvec{X}, \left(\myvec{b}\odot\myvec{s}^y\right)^\top \myvec{Y}\right) \\ &+ \lambda_x \norm{\myvec{s}^x}_0 + \lambda_y \norm{\myvec{s}^y}_0,
\end{split}
\end{equation*}
where $\odot$ denotes the Hadamard product and $\lambda_x$ and $\lambda_y$ are regularization parameters controlling the canonical vectors' sparsity.

By parameterizing the vectors $\myvec{s}^x$ and $\myvec{s}^y$ as the clamped Gaussian random variables $\mathbf{z}_x$ and $\mathbf{z}_y$ with means $\myvec{\mu}_x$ and $\myvec{\mu}_y$ respectively such that $z_d=\max(0,\min(1, \mu_d + \epsilon_d))$ 
where $\epsilon_d\sim\mathcal{N}(0,\sigma^2)$, and $\sigma$ is constant during training.
$\ell_0$-CCA goal can be written as

\begin{equation*}
\begin{split}
    \min_{\myvec{a}, \myvec{b}, \myvec{\mu}^x, \myvec{\mu}^y} \mathbb{E}  \bigg\{ &-\rho \left(\left(\myvec{a}\odot\mathbf{z}^x\right)^\top \myvec{X}, \left(\myvec{b}\odot\mathbf{z}^y\right)^\top \myvec{Y}\right) \\ &+ \lambda_x \norm{\mathbf{z}^x}_0 + \lambda_y \norm{\mathbf{z}^y}_0 \bigg \},
\end{split}
\end{equation*}
and the expectation over the gates boils down to
$
  \mathbb{E}_z\norm{\mathbf{z}}_0=\sum_{d=1}^D\mathbb{P}(z_d>0)=\sum_{d=1}^D\Phi\left(\frac{\mu_d}{\sigma}\right),
$
where $\Phi$ is the Gaussian PDF. $\ell_0$-CCA can be optimized using gradient descent.

$\ell_0$-CCA can be extended to non-linear projections as $\ell_0$-Deep CCA.
Given  ${\mathbf{f_\theta}_x}: \mathbb{R}^{D_x}\rightarrow \mathbb{R}^o$ and ${\mathbf{g_\theta}_y}: \mathbb{R}^{D_y}\rightarrow \mathbb{R}^o$, the optimization problem of $\ell_0$-Deep CCA is formulated as
\begin{equation*}
\begin{split}
    \min_{\boldsymbol{\theta}_x, \boldsymbol{\theta}_y, \boldsymbol{\mu}_x, \boldsymbol{\mu}_y} &-\rho\left(\mathbf{f}(\myvec{X}\odot\mathbf{z}_x), \mathbf{g}(\myvec{Y}\odot\mathbf{z}_y)\right) \\ &+\lambda_x\norm{\mathbf{z}_x}_0 + \lambda_y\norm{\mathbf{z}_y}_0.
\end{split}
\end{equation*}

\section{Method}
\label{sec:method}
To address the limitations of traditional temporal alignment methods in handling high-dimensional and sparse data, we propose Conditional Deep Canonical Time warping (CDCTW). This temporal alignment framework integrates a novel unsupervised adaptation of the stochastic gating layer.
Given the data matrices $\myvec{X}\in \mathbb{R}^{D_x\times N_x}$ and $\myvec{Y}\in \mathbb{R}^{D_y\times N_y}$ and the NNs ${\mathbf{f_\theta}_x}: \mathbb{R}^{D_x}\rightarrow \mathbb{R}^o$ and ${\mathbf{g_\theta}_y}: \mathbb{R}^{D_y}\rightarrow \mathbb{R}^o$, where $o$ is the embedding dimension and by using the notation introduced in Section \ref{sec:background}, the goal of CDCTW can be written as
\begin{equation*}
\begin{split}
    \min_{\boldsymbol{\theta}_x, \boldsymbol{\theta}_y} &-\rho\left(\mathbf{f}(\myvec{X}\odot\myvec{s}^x)^{(P_x)}, \mathbf{g}(\myvec{Y}\odot\myvec{s}^y)^{(P_y)}\right) \\ &+\lambda_x\norm{\myvec{s}^x}_0 + \lambda_y\norm{\myvec{s}^y}_0,
\end{split}
\end{equation*}
where $P_x$ and $P_y$ are the optimal alignment paths calculated using DTW. The optimization of this model requires optimizing the network and calculating DTW iteratively.

Since the data is time-dependent the selected features should also be different across samples. We achieve temporality in the gates by conditioning each input with temporal data related to the input. This is done by modeling the feature selection vectors $\myvec{s}^x$ and $\myvec{s}^y$ as the clamped Gaussian random variables $\mathbf{z}_x$ and $\mathbf{z}_y$ with means $\myvec{\mu}_x$ and $\myvec{\mu}_y$ respectively, where $\myvec{\mu}_x$ and $\myvec{\mu}_y$ are the output of the hypernetworks \cite{ha2016hypernetworks}
$\boldsymbol{\phi}$, $\boldsymbol{\psi}$ named "gating networks". The input to these gating networks are the data matrices $\mathcal{T}_x\in\mathbb{R}^{D_{\mathcal{T}x}\times N_x}$ and $\mathcal{T}_y\in\mathbb{R}^{D_{\mathcal{T}y}\times N_y}$ respectively. This implies the following gating vectors $\mathbf{z}_d=\max(0,\min(1, \mu_d + \epsilon_d))$ where $
    \myvec{\mu}_x = \boldsymbol{\phi}\left(\mathcal{T}_x\right), \;\;\; \;\myvec{\mu}_y = \boldsymbol{\psi}\left(\mathcal{T}_y\right)$
and $\epsilon_d\sim\mathcal{N}(0,\sigma^2)$, such that $\sigma$ is constant during training.

The gating networks $\boldsymbol{\phi}_{\eta_x}: \mathbb{R}^{D_{\mathcal{T}x}}\rightarrow \mathbb{R}^{D_x}$ and $\boldsymbol{\psi}_{\eta_y}: \mathbb{R}^{D_{\mathcal{T}y}}\rightarrow \mathbb{R}^{D_y}$ can be of any architecture but its output layer must be of the same number of features as its respective data matrices. We also impose restrictions on the activation function $\alpha(v)$ of the output layer of the gating functions. Since we wish the gates to converge to either 1's or 0's when clamped $\alpha(v)$ must satisfy $\lim_{v\to-\infty} \alpha(v)=0-\Delta$, $\lim_{v\to\infty} \alpha(v)=1+\Delta$, and it's derivative must satisfy $\lim_{v\to\pm\infty}\frac{d\alpha}{dv}=0$, $\forall v: \frac{d\alpha}{dv} > 0$.
The conditions on the activation guarantee a symmetric (fair) distribution around the clamping function, while the conditions on its derivative assure the convergence of the gates. The activation function we have used is $\alpha(v) = \tanh(v)+0.5$.

$\mathcal{T}_x$ and $\mathcal{T}_y$, the inputs to the hypernetworks, need to provide contextual-temporal information on $\myvec{X}$ and $\myvec{Y}$ respectively. The contextual information helps obtain the relevant features in a high dimensional space for each input, making the downstream task easier for the embedding networks. For tasks involving causal data, such as real-life audio or video, a viable choice for $\mathcal{T}_x$ and $\mathcal{T}_y$ would be one that involves some preprocessing on the difference between intra-view subsequent samples. This is true since by assuming causality we can further assume small change between subsequent samples. In our work, flow vector magnitude is used to predict the gates.

Using the temporal gates, CDCTW's goal can be written as
\begin{equation*}
\begin{split}
    \min_{\boldsymbol{\theta}_x, \boldsymbol{\theta}_y, \boldsymbol{\eta}_x, \boldsymbol{\eta}_y} &\mathbb{E}\bigg\{-\rho\left(\mathbf{f}\left(\mathbf{z}_x\odot\myvec{X}\right)^{(P_x)}, \mathbf{g}\left(\mathbf{z}_y\odot\myvec{Y}\right)^{(P_y)}\right) \\ &+\lambda_x\norm{\mathbf{z}_x}_0 + \lambda_y\norm{\mathbf{z}_y}_0\bigg\}.
\end{split}
\end{equation*}
The first part of the loss function assures a maximally correlated subspace in which the sequences are temporally aligned, while the second part is responsible for the feature selection. Our suggested model architecture is depicted in Fig. \ref{fig:arch}.

\section{Experiments}
\label{sec:experiments}
This section presents the datasets used in our experiments, introduces other state-of-the-art time-warping algorithms, defines our evaluation metric, and presents our results. Our model is compared to Canonical-Soft DTW with and without temperature annealing (CSTW/ASTW) \cite{kawano2019canonical}, Deep Canonical DTW (DCTW) \cite{trigeorgis2016deep}, Canonical DTW (CTW) \cite{zhou2009canonical}, Generelized DTW \cite{deriso2022general}, and DTW with either raw or PCA. For all non-deterministic methods, we conducted each experiment 10 times and reported the mean and standard deviation. For each of the experiments, we embedded the data in $\mathbb{R}^2$. 

\subsection{Datasets}
\label{sec:datasets}
\begin{figure}[t!]
    \centering
    \includegraphics[width=0.45\textwidth]{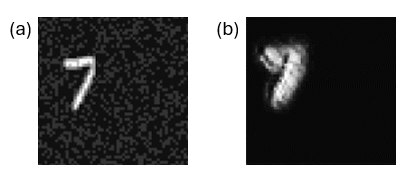}
    \caption{Moving MNIST: (a) A sample MNIST digit on a black background with added noise; (b) The corresponding learned gates.}
    \label{fig:moving_mnist}
\end{figure}

Our evaluation is separated into two parts. The first part is on a synthetic moving MNIST dataset, with results shown in Table \ref{tab:mnist_results}. We used samples from the MNIST dataset and moved them frame by frame along a shared random path, introducing varying levels of noise, as illustrated in Fig. \ref{fig:moving_mnist} (a). The second part of our evaluation, with results in Table \ref{tab:bench_results}, tests our model on four benchmark datasets.

\textbf{TCD-TIMIT} \cite{harte2015tcd}, an audio-visual dataset consisting of speakers reading different sentences. The task involves aligning two videos of different speakers.    
\textbf{MMI Facial Expression}, a visual dataset containing different subjects performing various facial expressions. The task was the alignment of smiles across different subjects.
\textbf{Weizmann} \cite{gorelick2007actions}, a visual dataset includes several subjects performing different human actions. The task is to align videos of different individuals walking. As this dataset lacks ground truth alignment paths, we inferred them by using the procedure suggested by \cite{trigeorgis2016deep}.    
\textbf{Synthetic Data}, as proposed in \cite{zhou2009canonical}, this synthetic dataset consists of three spatial dimensions and one temporal dimension.

\subsection{Evaluation}
\label{sec:evaluation}


We follow the alignment score used in \cite{trigeorgis2016deep, kawano2019canonical} which is defined as the ratio of correctly aligned frames/labels within each temporal phase to the total duration of the temporal phase across the aligned videos. This can be expressed as $\frac{|\kappa_x \cap \kappa_y|}{|\kappa_x \cup \kappa_y|}$, where $\kappa_x$ and $\kappa_y$ denote the sets of aligned frame indices obtained after warping the initial vector of annotations using the alignment matrices $P_x$ and $P_y$ determined through a temporal warping technique.


\subsection{Results}
\label{sec:results}

\begin{table}[htbp]
\caption{Moving MNIST Alignment Score}
\vspace{-0.5cm}
\begin{center}
\resizebox{\columnwidth}{!}{ 
\small 
\setlength{\tabcolsep}{3pt} 
\begin{tabular}{|c|c|c|c|}
\hline
Method & \parbox[c]{2cm}{\centering MNIST \\ ($\sigma = 0.102$)} & \parbox[c]{2cm}{\centering MNIST \\ ($\sigma = 0.421$)} & \parbox[c]{2cm}{\centering MNIST \\ ($\sigma = 0.737$)} \\
\hline
DTW & 0.000 & 0.000 & 0.000  \\
\hline
PCA DTW & 0.000 & 0.000 & 0.000  \\
\hline
GTW \cite{deriso2022general} & 0.148 & 0.001 & 0.002  \\
\hline
CTW \cite{zhou2009canonical} & 0.081$\pm$0.130 & 0.009$\pm$0.013 & 0.049$\pm$0.121 \\
\hline
DCTW \cite{trigeorgis2016deep} & 0.140$\pm$0.150 & 0.045$\pm$0.115 & 0.045$\pm$0.110 \\
\hline
CSTW \cite{kawano2019canonical} & 0.106$\pm$0.127 & 0.016$\pm$0.029 & 0.065$\pm$0.121 \\
\hline
ASTW \cite{kawano2019canonical} & 0.119$\pm$0.164 & 0.014$\pm$0.022 & 0.065$\pm$0.121 \\
\hline
CDCTW (Ours) & \textbf{0.247$\pm$0.174} & \textbf{0.211$\pm$0.145} & \textbf{0.249$\pm$0.138} \\
\hline
CDCTW (Ours) + ASTW \cite{kawano2019canonical} & 0.213$\pm$0.123 & 0.135$\pm$0.066 & 0.192$\pm$0.070 \\
\hline
\end{tabular}
}
\label{tab:mnist_results}
\end{center}
\end{table}
\begin{table}[htbp]
\caption{Benchmark Datasets Alignment Score}
\begin{center}
\resizebox{\columnwidth}{!}{ 
\small 
\setlength{\tabcolsep}{3pt} 
\begin{tabular}{|c|c|c|c|c|}
\hline
Method &  TIMIT  \cite{harte2015tcd} & MMI SMILE \cite{trigeorgis2016deep, kawano2019canonical} & \parbox[c]{2cm}{\centering Synthetic Data \cite{kawano2019canonical}} & \parbox[c]{2cm}{\centering Weizzman \cite{trigeorgis2016deep}} \\
\hline
DTW &  0.294 & 0.671 & 0.164 & 0.121 \\
\hline
PCA DTW &  0.132 & 0.686 & 0.115 & 0.067 \\
\hline
GTW \cite{deriso2022general} & 0.020 & 0.578 & 0.040 & 0.097 \\
\hline
CTW \cite{zhou2009canonical} & 0.252$\pm$0.150 & 0.684$\pm$0.165 & 0.020$\pm$0.020 & 0.174$\pm$0.077 \\
\hline
DCTW \cite{trigeorgis2016deep} & 0.351$\pm$0.230 & 0.596$\pm$0.162 & 0.019$\pm$0.053 & 0.255$\pm$0.189 \\
\hline
CSTW \cite{kawano2019canonical} & 0.364$\pm$0.308 & 0.548$\pm$0.172 & 0.070$\pm$0.095 & 0.183$\pm$0.120 \\
\hline
ASTW \cite{kawano2019canonical} & 0.365$\pm$0.307 & 0.547$\pm$0.172 & 0.054$\pm$0.092 & 0.179$\pm$0.116 \\
\hline
CDCTW (Ours) & 0.439$\pm$0.214 & \textbf{0.802$\pm$0.091} & \textbf{0.165$\pm$0.233} & \textbf{0.334$\pm$0.172} \\
\hline
CDCTW (Ours) + ASTW \cite{kawano2019canonical} & \textbf{0.476$\pm$0.254} & 0.800$\pm$0.081 & 0.125$\pm$0.217 & 0.257$\pm$0.170 \\
\hline
\end{tabular}
}
\label{tab:bench_results}\end{center}
\end{table}

The experimental results illustrate that our proposed method, CDCTW, outperforms existing alignment techniques. In the Moving MNIST alignment task in Tables \ref{tab:mnist_results}, CDCTW consistently outperforms other methods with significant margins, particularly as the noise level increases ($\sigma = 0.737$). This trend is similarly reflected in Table \ref{tab:bench_results}, where CDCTW demonstrates its effectiveness on four benchmark datasets. CDCTW achieves the highest alignment score in three out of four datasets, apart from TIMIT, where the highest score is a combination of CDCTW and ASTW. These results underscore the robustness and efficacy of CDCTW, particularly in scenarios involving high noise or complex, multi-view data.

\section{Conclusion}
\label{sec:conclusion}
This work introduces Conditional Deep Canonical Temporal Time Warping (CDCTW), a novel approach for improving temporal alignment in sparse and high-dimensional data. By integrating Conditional Stochastic Gates with Deep Canonical Correlation Analysis and temporal warping, CDCTW addresses limitations inherent in existing methods like DCTW, particularly in the presence of sparse and high-dimensional data. Our extensive experiments across various datasets demonstrate that CDCTW significantly outperforms previous alignment techniques, showcasing enhanced robustness and alignment accuracy. This advancement holds promise for diverse applications where precise temporal alignment is critical, paving the way for further innovations in the field of time series analysis.

\newpage
\label{sec:refs}
\bibliographystyle{IEEEbib}
\bibliography{bibliography}

\end{document}